# MMCAformer: Macro-Micro Cross-Attention Transformer for Traffic Speed Prediction with Microscopic Connected Vehicle Driving Behavior

Lei Han*, Mohamed Abdel-Aty, *Senior Member, IEEE,* Younggun Kim, Yang-Jun Joo, and Zubayer Islam, *Member, IEEE*

*Abstract*—Accurate speed prediction is crucial for proactive traffic management to enhance traffic efficiency and safety. Existing studies have primarily relied on aggregated, macroscopic traffic flow data to predict future traffic trends, whereas road traffic dynamics are also influenced by individual, microscopic human driving behaviors. Recent Connected Vehicle (CV) data provide rich driving behavior features, offering new opportunities to incorporate these behavioral insights into speed prediction. To this end, we proposed the Macro-Micro Cross-Attention Transformer (MMCAformer) to integrate CV data-based micro driving behavior features with macro traffic features for speed prediction. Specifically, MMCAformer employs self-attention to learn intrinsic dependencies in macro traffic flow and cross-attention to capture spatiotemporal interplays between macro traffic status and micro driving behavior. MMCAformer is optimized with a Student-t Negative Log-likelihood Loss to provide point-wise speed prediction and estimate uncertainty. Experiments on four Florida freeways demonstrate the superior performance of the proposed MMCAformer than baselines. Compared with only using macro features, introducing micro driving behavior features not only enhances prediction accuracy (e.g., overall RMSE, MAE, and MAPE reduced by 9.0%, 6.9%, and 10.2%, respectively) but also shrinks model prediction uncertainty (e.g., mean predictive intervals decreased by 10.1-24.0 % across the four freeways). Results reveal that hard braking and acceleration frequencies emerge as the most influential features. Such improvements are more pronounced under congested, low-speed traffic conditions.

*Index Terms*— Traffic Speed Prediction; Microscopic Driving Behavior; Connected Vehicle Data; Transformer Model; Uncertainty Estimation.

## I. INTRODUCTION

TRAFFIC speed prediction is a fundamental component of intelligent transportation systems. Accurate speed prediction enables traffic managers to implement proactive traffic controls, mitigating potential congestion and crashes and thereby enhancing traffic efficiency [1]. Utilizing predicted speed information, human travelers and intelligent vehicles can optimize their departure times and route choices to improve their mobility [2, 3]. Over the past decades, considerable research efforts have been devoted to this field, yielding significant improvements in both prediction accuracy and methodological development [1, 4-10].

In freeway environments, traffic speed is impacted by both macro-level traffic status (e.g., road volume) and micro-level driving behavior (e.g., hard braking) [11-13]. From a macro perspective, the future speed of is largely determined by the current traffic conditions as well as the states of surrounding segments. Existing studies have typically relied on aggregated macro-level traffic features (e.g., average speed and volume) and predict speed based on spatiotemporal dependencies [6-9]. At the micro level, road traffic dynamics are largely influenced by individual human driving behavior and their interaction along segments, which cannot be observed by traditional fixed sensors [11, 14]. As illustrated in **Fig. *1***, when a vehicle executes a hard braking, it not only reduces its own speed but also induces a deceleration wave that propagates to following vehicles, consequently leading to a future speed drop and congestion. Prior studies have shown that even a slight braking maneuver may disrupt traffic flow and trigger a stop-and-go wave [15-17]. Overall, the majority of existing speed prediction studies have focused on capturing macro-level traffic correlations. While a few studies explored the impacts of micro driving behaviors on traffic speed [11, 12], these efforts have largely relied on simulation data at specific locations. To date, integrating fine-grained behavioral information for real-world, large-scale road network speed prediction remains an open yet valuable challenge.

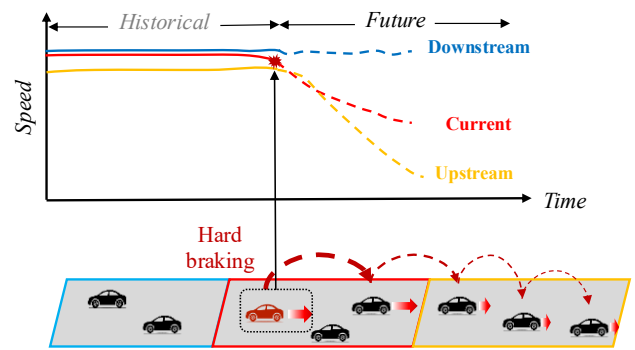

**Fig. 1.** Impact of hard braking on segment speeds.

This work was supported in part by the Florida Department of Transportation. *(Corresponding author: Lei Han).*

Lei Han, Mohamed Abdel-Aty, Younggun Kim, Yang-Jun Joo, and Zubayer Islam are with the Smart and Safe Transportation lab at the Department of Civil, Environmental and Construction Engineering, University of Central Florida, USA (e-mail: le966091@ucf.edu; m.aty@ucf.edu; younggun.kim@ucf.edu; yangjun.joo@ucf.edu; zubayer.islam@ucf.edu )

With the advances of emerging driving status monitoring and vehicle connection techniques, novel Connected Vehicle (CV) data can be flexibly acquired from diverse equipment (e.g., on-board units and smartphones). Unlike fixed sensors that provide traffic information at sparse and discrete points, CV trajectories span full road segments to capture continuous traffic dynamics along the roadway, resulting in a more comprehensive representation of traffic flow status [18, 19]. More importantly, CV data typically have high temporal resolution (1-3s updating), precise positioning (decimeter-level), and rich kinematics information (e.g., speed, acceleration) [20-23]. These high-resolution data enable the extraction of microscopic driving behavior features (e.g., hard acceleration and braking), which serve as critical indicators of traffic operations and potential traffic fluctuations [11, 14, 24]. Compared to the highly aggregated sensor-based traffic data, leveraging CV-derived microscopic driving behavior features into traffic speed prediction holds significant potential for effectively capturing traffic dynamics and enhancing prediction performance.

Given these advantages, this study aims to extract micro-level driving behavior measures from CV data and investigate their benefit for traffic speed prediction. However, introducing such driving behavior presents a set of challenges. First, recent studies have primarily focused on modeling spatial-temporal dependencies of macro traffic flow features with deep learning (DL) models (e.g., Spatial-Temporal Graph Neural Networks (STGNNs) [6, 25, 26] and Transformer models [7, 10, 27]). In contrast, micro-level driving behaviors are inherently discrete and event-driven to exhibit different spatial-temporal patterns compared to aggregated traffic flow features. This disparity makes it inefficient to directly incorporate these behavioral features into existing models without structural adaptations. Second, complex interactions and dependencies exist between macro traffic conditions and micro driving behaviors [11, 17]. On one hand, macro traffic states influence individual driving behaviors. For instance, hard acceleration and deceleration maneuvers are less likely to occur under free-flow conditions but become more frequent in congested traffic. On the other hand, driving behaviors also show heterogeneous impacts on traffic conditions. Frequent deceleration events, for example, can significantly reduce traffic speed in free-flow scenarios, but have less impact once traffic is already under congestion. How to effectively capture the bidirectional interactions between macro- and micro-level features poses a key challenge for developing accurate and behavior-aware traffic speed prediction frameworks.

To address the aforementioned research gaps, we propose a Macro-Micro Cross-Attention Transformer (MMCAformer) to effectively integrate both macro traffic flow features and micro driving behavior features for traffic speed prediction. Rather than utilizing simulated data at specific locations, this study collected real-world CV data and extracted both macro traffic flow and micro driving behavior features for the freeway network-level traffic speed forecasting. The main contributions of this paper include:

1) Multiple micro-level driving behavior features (e.g., speed volatility, hard acceleration/deceleration) are extracted for enhancing traffic speed prediction.
2) A cross-attention mechanism is designed in the MMCAformer to learn the spatiotemporal interplay between macro traffic states and micro driving behavior.
3) Student-t Negative Log-likelihood Loss is applied to address heavy-tailed prediction errors, enabling both point-wise speed prediction and uncertainty estimation.
4) Extensive experiments are conducted to demonstrate the SOTA performance of MMCAformer, highlighting the benefits of micro-level features to improve traffic speed prediction accuracy and reliability.

The paper is organized into seven sections. Section II presents the literature review, followed by the data preparation described in Section III. Section IV shows the details of the proposed MMCAformer, and Section V illustrates the experiment results. Finally, the discussions and conclusions are presented in Section VI and Section VII, respectively.

II. LITERATURE REVIEW

*A. Traffic Speed Prediction Data*

Traffic speed prediction studies have generally utilized two types of data sources: fixed sensor data and probe vehicle data. The majority of existing studies has focused on fixed sensor data collected from fixed loop detectors and microwave sensors for traffic speed prediction [2, 6-8, 25-28]. The most popular benchmarks are the open-source PeMS series (e.g., PeMSD4/7/8 and PeMSD-BAY) and METR-LA datasets. They were collected from loop detectors on several California freeways, including average speed, volume, and occupancy at 5min intervals. Traffic datasets from microwave sensors [1, 29] and Electronic Toll Collection [9] has been also used. Probe vehicle datasets are another important data source. Compared to fixed sensor data, these data are directly collected from vehicle GPS and can cover a wide variety of roadway types (e.g., urban arterials), thereby overcoming the spatial limitations of fixed sensor deployments [30]. Most of the existing studies collect data from taxi fleets to estimate the average speed of road segments [2, 31, 32]. For example, the SZ-Taxi dataset [8], one of the most widely used probe vehicle datasets, maps 10s GPS trajectories onto 156 major roads in Shenzhen, with average speeds aggregated at 15-min intervals. However, due to limitations in data collection, both fixed-sensor and probe-vehicle data are highly aggregated at the segment level, making it impossible to capture microscopic driving behavior features (e.g., hard acceleration and brake).

Recently, high-resolution CV trajectory data became more accessible to provide individual driving behaviors such as hard acceleration and braking [16, 35]. Several studies attempted to explore CV-based driving behavior features for various traffic prediction tasks. Most of these efforts focused on traffic safety, where risky driving behaviors were extracted to predict crash frequencies [24, 36] and real-time crash risks [23, 37]. In traffic speed analysis, [11] employed motorway CV data and found a mutual causality relationship between lane changing and congestion. While this work provides empirical evidence linking micro driving behaviors with macro traffic flow, it does not address speed prediction. [12] proposed a traffic state prediction model by fusing vehicles' acceleration and deceleration with detector data. However, their work relied on simulated CV data with high penetration rates of 10–50%, which is unrealistic at the current stage. Therefore, how to effectively integrate CV-derived microscopic driving behavior features into traffic speed prediction frameworks has not been studied and still needs further investigation.

*B. Traffic Speed Prediction Methods*

Existing traffic speed prediction methods can generally be divided into three categories: statistical methods, machine learning (ML) algorithms, and DL models. Statistical methods primarily include historical average (HA) [36], Kalman filters (KF) [37], autoregressive integrated moving average (ARIMA), and their variants [40, 41]. However, these methods rely on strict theoretical assumptions, making it difficult to accurately predict complex traffic speed changes. ML algorithms alleviate such issues by processing high-dimensional information and extracting nonlinear relationships in traffic flow data. Approaches such as support vector machine (SVM) [40], gradient boosting decision tree (GBDT) [41], hidden Markov model (HMM) [42], and artificial neural networks (ANN) [45, 46] have been adopted. Nevertheless, traffic data exhibits both temporal and spatial characteristics, while these methods only focus on temporal features and fail to model complex spatial dependencies.

In recent years, DL models have been developed to address the spatiotemporal modeling of traffic data. The STGNN has been proposed to simultaneously capture spatial and temporal dependencies in traffic flow data and achieved improved prediction performances [6, 47]. Subsequently, a variety of STGNN-based models (e.g., Graph WaveNet [46], ASTGCN [47], T-GCN [8], GMAN [48], STGIN [9], and SFGCN [49]) have been developed to explore spatiotemporal relationships from different perspectives. Recently, Transformer-based models have gained popularity and demonstrated strong performance, owing to their attention mechanisms that effectively learn the important spatiotemporal traffic dynamics [7, 10, 27, 52, 53]. For instance, STAEformer [50] leverages the powerful capabilities of a transformer with novel spatial, temporal, and adaptive embeddings to learn the spatial and temporal patterns in traffic data without explicit graph structures, providing an effective and powerful framework for improving speed prediction performance. Considering the outstanding performance of the Transformer-based model, we choose it as the base model for further improvement. However, existing models are primarily designed to capture the spatial-temporal relationships of macro traffic flow features (e.g., segment speed and volume), while the integration of micro-level driving behavior features into Transformer-based frameworks remains largely underexplored.

## III. DATA PREPARATION

*A. Connected Vehicle Dataset and Study Freeways*

In this study, StreetLight CV data were obtained from three Florida counties: Hillsborough, Orange, and Seminole. The Hillsborough dataset spans 21 days: Jan 3-13 and Jan 30-Feb 8, 2024. The Orange and Seminole dataset covers 45 days: May 1-31 and Oct 1-15, 2024. The CV data contains 3s-interval vehicle trajectory points, including journey Id, timestamp, GPS latitude and longitude, heading, and speed. On average, it comprises over 8.9 million CV points from 404,997 journeys per day. As the CV fleets are mainly collected from non-commercial vehicles, this dataset has a high penetration of 4-5% to better represent the road traffic. Within the study area, four freeways are selected as shown in **Fig. 2**. Three freeways (Hills_I4, Hills_I75, and Hills_I275) lie within Hillsborough County, while the fourth (OS_I4) spans both Orange and Seminole counties. The total length of these freeways is 226.5 miles. The four freeways were divided into a total of 565 segments for further speed analysis.

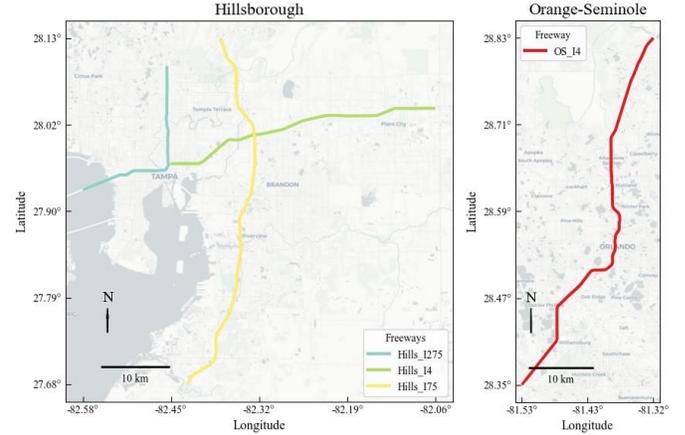

**Fig. 2.** Study freeways.

*B. Macro Traffic Flow and Micro Driving Behavior Feature Extraction*

Based on GPS locations, CV trajectories were first extracted and mapped to the corresponding freeway segments. Abnormal CV points with speeds exceeding 120 mph were then excluded. Vehicles that remained stationary within a segment for extended periods, likely due to mechanical failure, were also excluded to avoid biasing the estimation. The final processed dataset includes a total of about 2.5 million unique journeys with more than 354 million CV points. From this dataset, both macro traffic flow and micro driving behavior features were extracted for each segment:

*1) Macro traffic flow features*

Both segment speed and CV volume were calculated as macro traffic flow metrics. Specifically, individual CV average speeds were first calculated using their trajectory data and then aggregated to segment level to yield segment speed. CV volume is the total number of unique CV journey IDs traversing the segment to indicate the overall traffic volume.

*2) Micro driving behavior features*

To fully capture critical driving behaviors related to road traffic dynamics, CV speed volatility (SV) and the frequencies of acceleration and braking behaviors were extracted. Referring to the concept of speed/driving volatility [34], CV _SV is an effective indicator of traffic fluctuation, which can be defined as:

$$SV = \frac{1}{N}\sum_{i}^{N} STD_i, STD_i = \sqrt{\frac{1}{k_i-1}\sum_{ij}^{k_i}(S_{ij} - \bar{S}_t)^2} \quad (1)$$

where $STD_i$ is the standard deviation of speeds for the $i$-th CV trajectory; $k_i$ is the total number of trajectory points of the vehicle on the segment; $S_{ij}$ is the speed at the $j$-th trajectory point and $\bar{S}_t$ is its mean speed over the time interval.

To identify the acceleration and braking behaviors, the acceleration for the $i$-th CV at $j$-th observation is calculated:

$$ACC_{ij} = \frac{\Delta S_{ij}}{\Delta t_{ij}} = \frac{S_{ij} - S_{i,j-1}}{t_{ij} - t_{i,j-1}} \quad (2)$$

where $t_{ij}$ and $t_{i,j-1}$ are the recording timestamps at the $j$-th and $(j-1)$-th points, respectively. Noting that acceleration here is calculated as the change in speed over 3s intervals, which is smaller than the instantaneous acceleration. Therefore, traditional thresholds for hard acceleration and braking are inappropriate in this context. To this end, referring to existing studies [52, 53] and the acceleration distributions, the 85th and 97.5th percentiles were adopted as cutoffs to stratify driving behavior into three intensity levels: light, medium, and hard. Specifically, acceleration behaviors are classified as hard (>0.89 m/s²), medium (0.45 -0.89 m/s²), and light (<0.45 m/s²). Braking behaviors are classified as hard (>1.19 m/s²), medium (0.45-1.19 m/s²), and light (<0.45 m/s²).

Finally, two macro traffic flow features (segment speed and CV volume) and seven micro driving behavior features (CV_SV and frequencies of hard/medium/light acceleration and braking) were extracted at 5min intervals.

## IV. MACRO-MICRO CROSS-ATTENTION TRANSFORMER

We formulate the traffic speed prediction as a multivariate spatiotemporal forecasting task that jointly leverages both macro traffic flow and micro driving behavior features. Given $N$ road segments at each time step $t$, the macro-level input feature is represented by $X_t^{Macro} \in \mathbb{R}^{N \times d_M}$, and the micro-level input feature is denoted by $X_t^{micro} \in \mathbb{R}^{N \times d_m}$, where $d_M$ and $d_m$ are the respective feature dimensions. Thus, the traffic speed prediction task is $\hat{X}_{T+1:T+F} = f_\theta(X_{T-H+1:T}^{Macro}, X_{T-H+1:T}^{micro})$, $H$ and $F$ represent the length of historical and forecasting frames, respectively.

To extract spatial-temporal correlations across segment traffic and the interactions between macro and micro features, we designed our MMCAformer ($f_\theta$) with four key modules: Embedding Layer, Spatial Macro-Micro Attention Layer, Temporal Macro-Micro Attention Layer, and Speed Prediction Head, as shown in **Fig. 3**.

### A. Embedding Layer

To capture high-dimensional features of traffic dynamics, we first map the raw macro- and micro-level features into the latent space through Multilayer Perceptron (MLP) layers. Within historical data, macro-level traffic flow features $X_t^{Macro} \in \mathbb{R}^{N \times d_M}$ consist of segment speed and CV volume, resulting in an input dimension of $d_M = 2$. The micro-level features comprise seven indicators: CV_SV and frequencies of hard, medium, and light acceleration and braking, giving an input dimension of $d_m = 7$. For each time step $t \in \{T - H + 1, \cdots, T\}$, macro features and micro features are independently transformed via fully connected layers:

$$E_f^{Macro} = FC^{Macro}(X_{T-H+1:T}^{Macro}) \quad (3\text{-}1)$$
$$E_f^{micro} = FC^{micro}(X_{T-H+1:T}^{micro}) \quad (3\text{-}2)$$

where $E_f^{Macro}, E_f^{micro} \in \mathbb{R}^{H \times N \times d_f}$ with $d_f$ as the dimension of these embeddings.

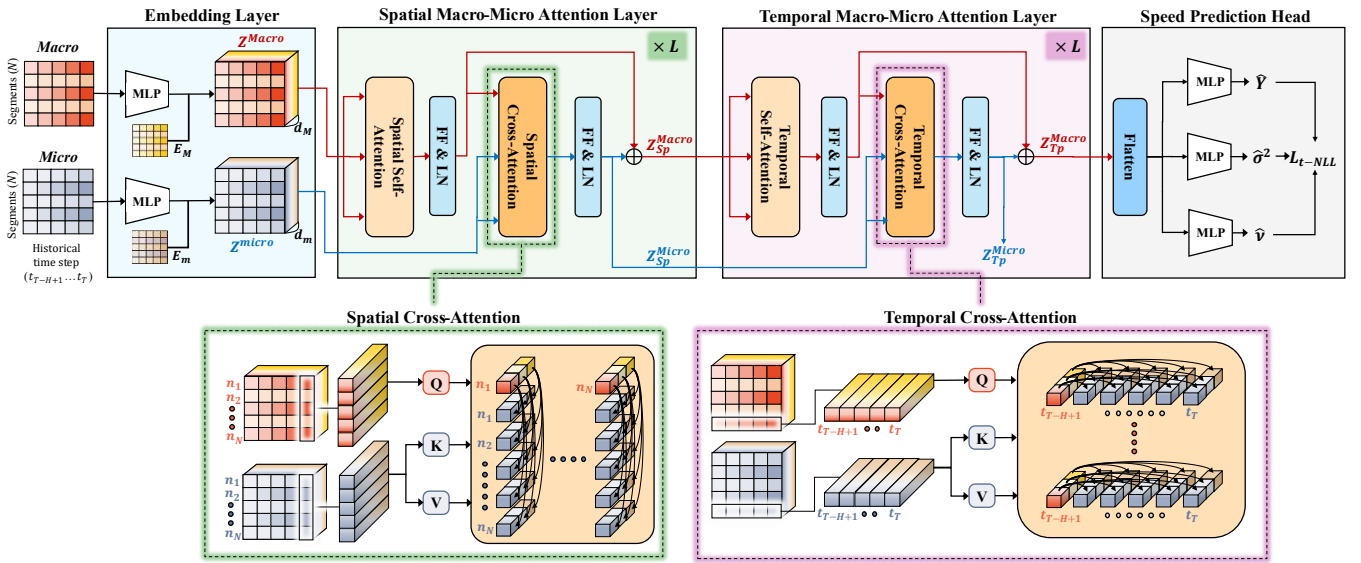

**FF & LN:** Feedforward and Layer Normalization; **MLP:** Multilayer Perceptron

**Fig. 3.** Model architecture of MMCAformer.

Inspired by [7, 50], the periodicity embedding and adaptive embedding are also integrated into the input embedding. For periodicity embedding, the Day-of-Week (DoW) and Time-of-Day (ToD) embedding are used to capture typical traffic patterns associated with weekly cycles (e.g., weekday vs. weekend) and daily periodicity (e.g., rush-hour congestion). Both DoW and ToD embeddings are retrieved across all spatial nodes for both macro and micro features to obtain $E_{DoW\_M/m} \in \mathbb{R}^{H \times N \times d_{Dow}}, E_{ToD\_M/m} \in \mathbb{R}^{H \times N \times d_{ToD}}$. Meanwhile, an adaptive embedding [50] is adopted as learnable spatiotemporal parameters $E_{adp\_M/m} \in \mathbb{R}^{H \times N \times d_a}$ for macro and micro features, respectively. They serve as position-dependent embeddings that adaptively encode spatiotemporal information for each time step and segment. Finally, the projected input features, DoW and ToD embeddings, and corresponding adaptive embeddings are concatenated to obtain the final macro and micro feature representations: $Z^{Macro/Micro} \in \mathbb{R}^{H \times N \times d_h}$, with $d_h = d_f + d_{Dow} + d_{ToD} + d_a$.

*B. Spatial Macro-Micro Attention Layer*

To capture spatial dependences within macro-level features (i.e., segment speed and CV volume), a spatial self-attention mechanism is first applied over segments for the embedded macro features $Z^{Macro}$:

$$A_{Sp}^{self} = Softmax\left(\frac{Q_{Sp}^{self}[K_{Sp}^{self}]^T}{\sqrt{d_h}}\right), \begin{cases} Q_{Sp}^{self} = Z^{Macro}W_Q^{self} \\ K_{Sp}^{self} = Z^{Macro}W_K^{self} \\ V_{Sp}^{self} = Z^{Macro}W_V^{self} \end{cases}$$
(4)

where $W_Q^{self}, W_K^{self}, W_V^{self} \in \mathbb{R}^{d_h \times d_h}$ are learnable weight matrices for self-attention. Here, $A_{Sp}^{self} \in \mathbb{R}^{H \times N \times N}$ captures spatial correlations across all $N$ segments: "*how the segment speed is influenced by its neighbors' macro traffic*". The output of spatial self-attention is computed as:

$$Z_{Sp}^{self} = A_{Sp}^{self} V_{Sp}^{self}$$
(5)

where $Z_{Sp}^{self} \in \mathbb{R}^{H \times N \times d_h}$, referred to as the spatial-attention macro features.

Next, spatial interactions among macro and micro features are modeled to enhance macro-level representations by integrating informative signals from driving behavior features. To this end, a spatial cross-attention mechanism is applied, enabling each macro node to attend to micro features at different road segments. Taking the spatial-attention macro features $Z_{Sp}^{self}$ and micro features $Z^{micro}$ as input, the attention score matrix is computed as:

$$A_{Sp}^{cros} = Softmax\left(\frac{Q_{Sp}^{cross}[K_{Sp}^{cross}]^T}{\sqrt{d_h}}\right), \begin{cases} Q_{Sp}^{cross} = Z_{Sp}^{self}W_Q^{cross} \\ K_{Sp}^{cross} = Z^{Micro}W_K^{cross} \\ V_{Sp}^{cross} = Z^{Micro}W_V^{cross} \end{cases}$$
(6)

where $W_Q^{cross}, W_K^{cross}, W_V^{cross} \in \mathbb{R}^{d_h \times d_h}$ are the learnable weight matrices for cross-attention [54]. The resulting cross-attention (CA) score matrix $A_{Sp}^{cros} \in \mathbb{R}^{H \times N \times N}$ captures the relevance between each macro node and micro nodes: "*how the segment speed is affected by the micro driving behavior features at its own and other segments*". The output of the spatial cross-attention layer is computed as:

$$Z_{Sp}^{cross} = A_{Sp}^{cros} V_{Sp}^{cross}$$
(7)

where $Z_{Sp}^{cross} \in \mathbb{R}^{H \times N \times d_h}$, referred to as the macro-micro spatial cross-attention features.

Finally, $Z_{Sp}^{cross}$ and $Z_{Sp}^{cross}$ are fused to produce the final spatial-enhanced macro features:

$$Z_{Sp}^{Macro} = LayerNorm(Z_{Sp}^{self} + Z_{Sp}^{cross})$$
(8)

Noting that the Spatial Macro-Micro Attention Layer is stacked $L$ times, enabling iterative interactions between macro and micro modalities at the spatial level. By integrating both self-attention and macro–micro cross-attention, this module captures intrinsic spatial dependencies among macro features while simultaneously incorporating the influences of micro driving behaviors.

*C. Temporal Macro-Micro Attention Layer*

The temporal macro-micro cross-attention layer is designed to model temporal dependencies across different time steps. This layer shares the same architecture as the spatial macro-micro cross attention layer, but the attention operation is applied along the temporal dimension.

First, temporal self-attention is applied to the spatial-enhanced macro features $Z_{Sp}^{Macro}$ to capture its temporal continuity and semantic structure, such as temporal changing trends in segment speed and CV volume over time. The temporal self-attention score matrix $A_{Tp}^{self} \in \mathbb{R}^{N \times H \times H}$ captures dependencies across all time steps: "*how the segment speed is influenced by its historical macro traffic conditions*". The output of the self-attention layer is a temporally contextualized macro representation $Z_{Tp}^{self} \in \mathbb{R}^{H \times N \times d_h}$.

Then, temporal cross-attention is performed to enhance the macro sequence with micro-level behavioral information, where the attention is computed between macro queries and micro keys/values over the temporal axis, as shown in **Fig. 3**. The temporal cross-attention score matrix $A_{Tp}^{cross} \in \mathbb{R}^{N \times H \times H}$ captures the relevance between macro and micro representations over time for each node: "*how the segment speed is affected by its historical driving behavior series*". The output of this module is denoted as $Z_{Tp}^{cross} \in \mathbb{R}^{H \times N \times d_h}$, referring to as micro-guided temporal features. Finally, macro and micro outputs are fused as follows:

$$Z_{Tp}^{Macro} = LayerNorm(Z_{Tp}^{self} + Z_{Tp}^{cross})$$
(9)

Similarly, the Temporal Macro-Micro Attention Layer is also stacked $L$ times to capture long-range temporal dependencies and refine macro-micro interactions over time. Therefore, this module captures not only the intrinsic temporal dependencies among macro features but also their joint interactions with micro driving-behavior features.

> REPLACE THIS LINE WITH YOUR MANUSCRIPT ID NUMBER (DOUBLE-CLICK HERE TO EDIT) <                                                                6*D. Speed Prediction Head with Uncertainty Estimation*

The final prediction head takes the final temporally enriched macro representations $Z_{Tp}^{Macro}$ from the last temporal macro-micro cross attention layer to predict future speeds. Recently, researchers have begun emphasizing the quantification of traffic speed uncertainty, a critical component for evaluating model confidence and reliability in traffic management applications [2]. Therefore, we set our model to not only predict the point-wise speed but also the predictive variance (mean-variance estimation) [2, 55, 56]. Although existing studies commonly assume the prediction error follows a Gaussian distribution [57, 58], our empirical analysis reveals that speed prediction errors exhibit heavy-tailed characteristics (see Section VI). To this end, the Student-t distribution, an extension of the Gaussian distribution, is applied to better estimate prediction errors and their uncertainty. Specifically, three parallel fully connected layers are designed to generate the predicted mean $\hat{Y} \in \mathbb{R}^{N \times F}$, scale parameter $\hat{\sigma}^2 \in \mathbb{R}^{N \times F}$, and degrees-of-freedom $\hat{v} \in \mathbb{R}^{N \times F}$ of predicted speeds:

$$\begin{cases} \hat{Y} = FC_{mean}\left(Flatten(Z_{Tp}^{Macro})\right) \\ \hat{\sigma}^2 = Softplus(FC_{var}\left(Flatten(Z_{Tp}^{Macro})\right)) \\ \hat{v} = Softplus(FC_{df}\left(Flatten(Z_{Tp}^{Macro})\right)) + 2 \end{cases} \quad (10)$$

Thus, the proposed MMCAformer can be optimized by the Student-t Negative Log-likelihood Loss (t-NLL) defined as:

$$L_{t-NLL} = \frac{1}{NF} \sum_{n=1}^{N} \sum_{t=T+1}^{T+F} \left[ \frac{\hat{v}_{n,t}+1}{2} \log\left(1 + \frac{(Y_{n,t}-\hat{Y}_{n,t})^2}{\hat{v}_{n,t}\hat{\sigma}_{n,t}^2}\right) + \frac{1}{2} \log(\hat{v}_{n,t}\hat{\sigma}_{n,t}^2\pi) + \log \Gamma\left(\frac{\hat{v}_{n,t}}{2}\right) - \log \Gamma\left(\frac{\hat{v}_{n,t}+1}{2}\right) \right] \quad (11)$$

## V. EXPERIMENTS

*A. Experiment Setup*

The processed CV data was used to conduct experiments. Consistent with prior studies [1, 8], the model takes the past one hour observations (*H*=12) and to predict speed over the next one hour (*F*=12). To mitigate low CV samples overnight, only data from 06:00 to 22:00 each day were included. TABLE I shows the train and test datasets that are split by date to prevent potential information leakage. Before model training, the Max-Min method was used to normalize the dataset to the range 0-1. During the testing, results were scaled back to the normal range and compared with the ground truth. To find the best hyperparameters of the proposed MMCAformer, a grid search was used and evaluated using 10% of the train dataset as validation. TABLE II summarizes the final model hyperparameter setting of the MMCAformer.

To demonstrate the effectiveness of our proposed model, seven existing prediction models were chosen as baselines: Seq2SeqLSTM in [29]; DCRNN in [45]; STGCN in [6]; Graph-WaveNet in [46]; ASTGCN in [47]; iTransformer in [10]; and SATEformer in [50]. In terms of evaluation metrics, three metrics are used to determine the difference between the observed speed $Y_{n,t}$ and the predicted speed $\hat{Y}_{n,t}$: the Root Mean Square Error (RMSE), Mean Absolute Error (MAE), and Mean Absolute Percentage Error (MAPE):

$$RMSE = \sqrt{\frac{1}{NF} \sum_{n=1}^{N} \sum_{t=1}^{F} (Y_{n,t} - \hat{Y}_{n,t})^2} \quad (12\text{-}1)$$

$$MAE = \frac{1}{NF} \sum_{n=1}^{N} \sum_{t=1}^{F} |Y_{n,t} - \hat{Y}_{n,t}| \quad (12\text{-}2)$$

$$MAPE = \frac{1}{NF} \sum_{n=1}^{N} \sum_{t=1}^{F} \frac{|Y_{n,t} - \hat{Y}_{n,t}|}{Y_{n,t}} \times 100\% \quad (12\text{-}3)$$

To ensure fair comparisons, we use the same training and test sets for all baselines. Each baseline was implemented according to its original specifications—either as described in the paper or in the official code—and then subjected to a grid search over its best model hyperparameters. We selected the hyperparameter set that yields the highest validation performance for final testing. During model training, every model was run for up to 100 epochs with a batch size of 32. An early-stop mechanism was used in all experiments, and the number of early-stop epochs is set to 10, matching the MMCAformer's settings. Although these baselines were originally designed for only macro traffic flow inputs (e.g., traffic speed and volume), we augmented their input vectors by concatenating micro driving behavior features with the macro flow features, thereby allowing them to exploit the same micro-level driving behavior information. Furthermore, the MMCAformer and all baselines were implemented in the PyTorch framework under the same hardware environment (NVIDIA RTX 3060 GPU), thereby ensuring consistency in running conditions.

TABLE I
TRAINING AND TEST DATA SPLITTING

| Models | Train | Test |
|---|---|---|
| Hills_I4, I75, I275 | Jan 3-11 & Jan 30-Feb 5 (16 days) | Jan 12-13 & Feb 6-8 (5 days) |
| OS_I4 | May 1-25 & Oct 1-7 (32 days) | May 26-31 & Oct 11-15 (11 days) |

TABLE II
THE OPTIMAL MMCAFORMER HYPERPARAMETER SETTING

| | Hyperparameter | Value |
|---|---|---|
| Feature Embedding | Input embedding dimension $d_f$ | 24 |
| | Tod embedding dimension $d_{Dow}$ | 2 |
| | Dow embedding dimension $d_{ToD}$ | 2 |
| | Adaptive embedding dimension $d_{ToD}$ | 80 |
| Spatial and Temporal layer | Stacked number *L* | 3 |
| | Number of blocks in Attention | 3 |
| | Number of heads in Attention | 4 |
| | MLP hidden nodes | 12 |
| Model training | Batch size | 32 |
| | Dropout | 0.1 |
| | Epochs | 100 |
| | Learning rate | 5e-4 |
| | Optimizer | Adam |

TABLE III
OVERALL MODEL PERFORMANCE COMPARISON ON DIFFERENT FREEWAYS

| Models | Hills_I4 | | | Hills_I75 | | | Hills_I275 | | | OS_I4 | | |
|---|---|---|---|---|---|---|---|---|---|---|---|---|
| | RMSE | MAE | MAPE | RMSE | MAE | MAPE | RMSE | MAE | MAPE | RMSE | MAE | MAPE |
| Seq2SeqLSTM | 8.61 | 5.08 | 13.16 | 8.35 | 4.41 | 12.12 | 9.18 | 5.44 | 18.45 | 7.74 | 4.47 | 20.04 |
| DCRNN | 8.46 | 4.95 | 13.31 | 6.68 | 3.63 | 9.04 | 7.71 | 4.85 | 15.98 | 7.21 | 4.14 | 17.36 |
| STGCN | 8.06 | 4.75 | 12.08 | 6.94 | 3.69 | 9.76 | 7.91 | 5.07 | 16.47 | 6.80 | 3.86 | 15.21 |
| GraphWaveNet | 8.03 | 4.73 | 11.82 | 6.58 | 3.63 | 9.54 | 7.69 | 4.69 | 15.16 | 6.60 | 3.90 | 15.77 |
| ASTGCN | 8.24 | 4.83 | 12.64 | 7.62 | 3.99 | 10.99 | 8.23 | 5.17 | 17.15 | 7.36 | 4.23 | 18.33 |
| iTransformer | 8.65 | 4.97 | 13.02 | 7.17 | 3.77 | 9.35 | 8.93 | 5.53 | 17.92 | 7.50 | 4.18 | 17.32 |
| SATEformer | 8.43 | 4.75 | 12.21 | 6.62 | 3.62 | 9.19 | 7.72 | 4.67 | 15.58 | 6.60 | 3.74 | 16.65 |
| **MMCAformer** | **7.49** | **4.46** | **11.41** | **5.78** | **3.31** | **7.34** | **6.71** | **4.33** | **14.07** | **5.86** | **3.43** | **14.13** |
| *(Decrease\*)* | *(6.7%)* | *(5.9%)* | *(3.5%)* | *(12.2%)* | *(8.6%)* | *(18.8%)* | *(12.8%)* | *(7.7%)* | *(7.1%)* | *(11.3%)* | *(11.9%)* | *(7.1%)* |

*: Compared to the best performance in baselines.

## B. Predicting Performance Comparison

TABLE III presents the overall prediction performance of the baselines and the proposed MMCAformer. The top-performing method is highlighted in **bold**, while the runner-up is underlined. Among the baselines, Seq2SeqLSTM and iTransformer have the weakest performances with high prediction errors. They only consider temporal correlations while neglecting spatial correlations between road segments. In contrast, GraphWaveNet and SATEformer have relatively optimal performances due to their powerful spatiotemporal modeling ability via graph network and Transformer architectures. Overall, MMCAformer performs well among the four freeways to achieve the lowest RMSEs, MAEs, and MAPEs. Consequently, it reduces the overall speed prediction errors by **6.7-12.8%** in RMSE, **5.9-11.9%** in MAE, and **3.5-18.8%** in MAPE compared with the best-performing baselines. Given that the baseline models were designed to capture the spatial-temporal relationships among macro features (e.g., speed and volume), they simply concatenate macro traffic-flow and micro driving behavior features during the modeling. In contrast, MMCAformer embeds the macro and micro features separately and leverages cross-attention layers to more effectively model their co-influences at both spatial and temporal dimensions.

## C. Ablation Studies

To validate the effectiveness of each essential component of the proposed MMCAformer, we conducted ablation studies with three variants of our model:

(1) **w/o temporal layers.** The Temporal Macro-Micro Attention Layers were removed.
(2) **w/o spatial layers.** The Spatial Macro-Micro Attention Layers were removed.
(3) **w/o cross-attention.** The cross-attention layers at both spatial and temporal layers were replaced by concatenation operation.

TABLE IV presents the ablation study results, showing that the removal of any component degrades model performances. In particular, variants without either temporal or spatial layers exhibit a substantial increase in prediction errors. This finding aligns with findings of existing studies [3, 27, 44], as accurate speed forecasting highly depends on capturing spatiotemporal correlations, which links a target segment's speed not only with its own historical speeds (temporal) but also with its neighboring segments (spatial). The proposed cross-attention layer is also critical for speed prediction, as variants that omit it yield higher prediction errors than the full MMCAformer. Compared with simple feature concatenation, the complete MMCAformer with cross-attention layers achieves notable improvements, reducing RMSE, MAE, and MAPE by **9.6%, 6.7%,** and **8.7%** at the average level, respectively. These results highlight the essential role of cross-attention in capturing the interplay between macro and micro features.

TABLE IV
ABLATION STUDY RESULTS

| Freeway | Metric | w/o temporal layers | w/o spatial layers | w/o cross-attention | All |
|---|---|---|---|---|---|
| Hills_I4 | RMSE | 8.23 | 8.61 | 8.02 | **7.49** |
| | MAE | 4.87 | 5.17 | 4.65 | **4.46** |
| | MAPE | 12.99 | 13.62 | 12.11 | **11.41** |
| Hills_I75 | RMSE | 6.93 | 6.74 | 6.49 | **5.78** |
| | MAE | 3.78 | 3.71 | 3.64 | **3.31** |
| | MAPE | 10.27 | 9.43 | 8.71 | **7.34** |
| Hills_I275 | RMSE | 7.92 | 8.22 | 7.52 | **6.71** |
| | MAE | 4.90 | 5.09 | 4.63 | **4.33** |
| | MAPE | 17.00 | 16.65 | 15.26 | **14.07** |
| OS_I4 | RMSE | 6.75 | 6.86 | 6.55 | **5.86** |
| | MAE | 3.77 | 3.88 | 3.74 | **3.43** |
| | MAPE | 17.21 | 16.45 | 15.37 | **14.13** |
| AVG | RMSE | 7.46 | 7.61 | 7.14 | **6.46** |
| | MAE | 4.33 | 4.46 | 4.16 | **3.88** |
| | MAPE | 14.37 | 14.04 | 12.86 | **11.74** |

*D. Effects of Micro Driving Behavior Features*

To investigate the benefits of introducing micro driving behavior features, further experiments were conducted to compare the model performances without (*w/o*) and with (*w*) such features, and the results are shown in TABLE V. Compared with the variant omitting micro driving behavior features, integrating such critical features substantially reduces prediction errors. Specifically, overall RMSE, MAE, and MAPE decrease by **9.0–14.0%, 6.9–10.1%,** and **10.2–20.5%,** respectively. Results indicate that introducing driving behavior features enables models to capture more traffic dynamics (e.g., traffic fluctuations) that cannot be reflected by the macro-level variables, thereby improving prediction accuracy.

Such improvements are also examined at different traffic statuses: extremely congested (0–20 mph), congested (20–40 mph), medium traffic (40–60 mph), and free-flow (>60 mph). Results show that the improvements are more pronounced in the extremely congested (0-20mph) and congested (20-40mph) regimes. For instance, on the freeway Hills_I75, RMSE reduces by 17.7% and MAE by 25.9%. The underlying reason may be that the micro behaviors (e.g., hard braking, speed volatility) are more likely to occur during the transition from free flow to congested traffic. Such micro information emerges as early signals for indicating imminent changes in speed (e.g., traffic slowdown), enabling the model to more accurately predict congested condition. Since speed prediction under congestion has been seen as a challenging task [7], integration of micro driving behavior features appears to be a promising solution to mitigate this issue.

To further investigate the impact of different micro driving behaviors, the effects of removing each micro driving behavior feature on model performance are analyzed in **Fig. 4**.

The results show that removing any micro driving behavior feature would increase prediction errors. Among these, hard acceleration and hard braking frequencies show relatively high impacts on the traffic speed prediction, as the MAE increases significantly when these features are removed. It indicates that these extreme driving maneuvers have significant impacts on the future traffic evolution as they may disrupt smooth flow and trigger traffic fluctuations [59, 60]. As the intensity levels of driving behavior decrease (e.g., medium and light braking), the impacts on prediction errors become smaller. Overall, the impacts of braking behaviors are relatively higher than accelerations, as frequent braking behaviors are more likely to cause traffic flow oscillations [17]. Additionally, the CV speed volatility also has a high impact, as it directly reflects traffic fluctuations on road segments, which helps the model predict potential traffic instability and speed drops.

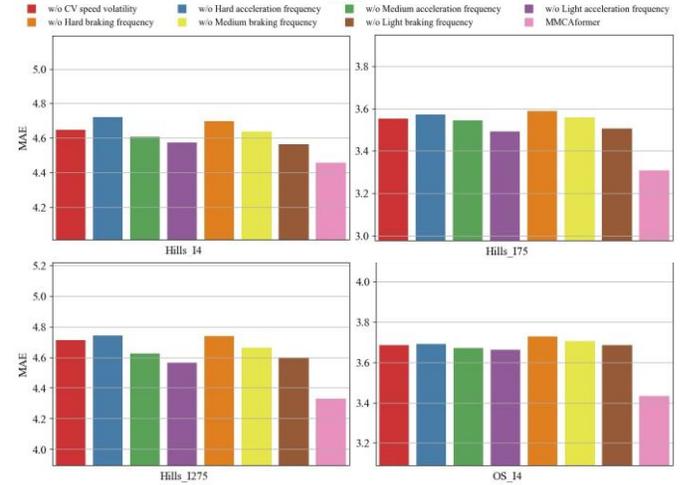

**Fig. 4.** Impact of different micro driving behavior features.

TABLE V
MODEL PERFORMANCES WITHOUT AND WITH MICRO DRIVING BEHAVIOR FEATURES

| Speed bins | Micro features | Hills_I4 | | | Hills_I75 | | | Hills_I275 | | | OS_I4 | | |
|---|---|---|---|---|---|---|---|---|---|---|---|---|---|
| | | RMSE | MAE | MAPE | RMSE | MAE | MAPE | RMSE | MAE | MAPE | RMSE | MAE | MAPE |
| 0-20 mph | *w/o* | 21.74 | 16.43 | 129.78 | 25.24 | 19.19 | 190.51 | 15.21 | 10.50 | 115.42 | 21.10 | 16.15 | 418.83 |
| | *w* | **19.88** | **14.58** | **113.78** | **20.78** | **14.22** | **133.85** | **14.14** | **10.35** | **110.32** | **18.07** | **13.62** | **329.52** |
| | *decrease* | *8.5%* | *11.3%* | *12.3%* | *17.7%* | *25.9%* | *29.7%* | *7.0%* | *1.5%* | *4.4%* | *14.4%* | *15.7%* | *21.3%* |
| 20-40 mph | *w/o* | 17.58 | 13.66 | 48.23 | 16.85 | 13.10 | 46.79 | 10.41 | 7.76 | 27.45 | 15.02 | 11.19 | 40.22 |
| | *w* | **16.77** | **13.17** | **46.35** | **15.94** | **12.20** | **43.07** | **9.40** | **7.14** | **25.60** | **13.76** | **10.53** | **37.80** |
| | *decrease* | *4.6%* | *3.6%* | *3.9%* | *5.4%* | *6.8%* | *8.0%* | *9.8%* | *8.0%* | *6.7%* | *8.4%* | *5.9%* | *6.0%* |
| 40-60 mph | *w/o* | 10.16 | 7.67 | 15.17 | **9.67** | **7.12** | **14.14** | 7.12 | 4.98 | 9.38 | 6.90 | 4.86 | 9.23 |
| | *w* | **9.73** | **7.46** | **14.79** | 10.59 | 8.09 | 16.02 | **6.17** | **4.34** | **8.18** | **6.32** | **4.49** | **8.53** |
| | *decrease* | *4.3%* | *2.7%* | *2.5%* | - | - | - | *13.4%* | *12.9%* | *12.8%* | *8.5%* | *7.6%* | *7.6%* |
| >60 mph | *w/o* | 5.20 | 3.24 | 4.72 | 4.76 | 2.91 | 4.00 | 5.68 | 3.68 | 5.60 | 4.11 | 2.64 | 3.91 |
| | *w* | **4.48** | **2.98** | **4.34** | **4.07** | **2.65** | **3.69** | **4.82** | **3.09** | **4.70** | **3.64** | **2.41** | **3.57** |
| | *decrease* | *13.8%* | *8.0%* | *8.1%* | *14.5%* | *8.8%* | *7.8%* | *15.2%* | *16.2%* | *16.1%* | *11.5%* | *8.5%* | *8.7%* |
| Overall | *w/o* | 8.23 | 4.79 | 12.91 | 6.64 | 3.64 | 9.23 | 7.80 | 4.82 | 15.68 | 6.52 | 3.75 | 15.88 |
| | *w* | **7.49** | **4.46** | **11.41** | **5.78** | **3.31** | **7.34** | **6.71** | **4.33** | **14.07** | **5.86** | **3.43** | **14.13** |
| | *decrease* | *9.0%* | *6.9%* | *11.7%* | *13.0%* | *9.1%* | *20.5%* | *14.0%* | *10.1%* | *10.2%* | *10.2%* | *8.5%* | *11.0%* |

*E. Uncertainty Evaluation*

To evaluate the uncertainty results, we compared models *w* and *w/o* micro features using two commonly used metrics: Mean Predictive Interval Widths (MPIW) and Prediction Interval Coverage Probability (PICP). Given the significance level of $\alpha$, the model can predict the predicted upper bound ($\hat{Y}_{U,(ij)}$) and lower bound ($\hat{Y}_{L,(ij)}$). Then the MPIW and PICP can be calculated:

$$MPIW = \frac{1}{DT}\sum_{i=1}^{D}\sum_{j=1}^{T}\hat{Y}_{U,(ij)} - \hat{Y}_{L,(ij)} \quad (13)$$

$$PICP = \frac{1}{DT}\sum_{i=1}^{D}\sum_{j=1}^{T}k_{ij}, k_{ij} = \begin{cases} 1, & if\ \hat{Y}_{L,(i,j)} \leq Y_{ij} \leq \hat{Y}_{U,(i,j)} \\ 0, else \end{cases} \quad (14)$$

Ideally, PICP should be close to the pre-set confidence level. While a smaller MPIW reflects tighter intervals and lower uncertainty, and therefore more reliable predictions.

Given the significance level of 90%, TABLE VI shows the uncertainty quantification results. Models with micro features consistently achieve better PICP values (0.88-0.91) to be close to 0.9, demonstrating better prediction intervals with the intended coverage, neither underestimating nor overestimating uncertainty. Moreover, models with micro features always produce narrower MPIWs across all prediction horizons, indicating lower uncertainty. For instance, as the prediction time increases, the MPIW of model without micro features rises sharply (e.g., from 14.8 to 27.1 on Hills_I4), whereas the model with micro features shows a more gradual increase (e.g., 13.2 to 17.0 on Hills_I4). On average, MPIW decreased by **24.0 %** on Hills_I4, **24.0 %** on Hills_I75, **16.4 %** on Hills_I275, and **10.1 %** on OS_I4, respectively.

**Fig. 5** presents an example to compare the uncertainty prediction results of an arbitrarily selected segment on OS_I4. **Fig. 5**(a) shows the prediction results of model with only macro features, while **Fig. 5**(b) shows the results for the model integrating micro driving behavior features. **Fig. 5**(c) gives the temporal evolution of three key driving behavior features: CV speed volatility, hard acceleration frequency, and hard braking frequency. During free-flow traffic status (e.g., 07:00-9:00 and 15:00-22:00), both models yield similarly narrow prediction intervals at a 90% significance level, reflecting the ease of forecasting stable speeds in uncongested traffic [4, 7]. Since critical driving behaviors (e.g., hard braking and accelerations) are minimal in such situations, micro features offer limited benefits. However, once the traffic congestion occurs (i.e., 11:30-13:30 on May 26th, and 11:00-13:30 on May 27th), the prediction uncertainties of model w/o micro features become very high (e.g., average MPIW are 25.5mph and 28.1mph). By contrast, the model integrating driving behavior features keeps relatively narrower prediction intervals (e.g., average MPIW of 15.1mph and 17.4mph). These reductions coincide with pronounced spikes in CV speed volatility and hard braking/acceleration frequency (**Fig. 5**(c)), indicating that such micro driving behavior features enable the model to effectively capture the low-speed traffic dynamics and therefore provide a more stable and confident prediction.

Therefore, these results illustrate that incorporating micro driving behavior features not only sharpens point forecasts but also significantly reduces prediction uncertainty, especially under congested, low-speed traffic conditions.

TABLE VI
UNCERTAINTY EVALUATION RESULTS (SIG. LEVEL=90%)*.

| Predict time | Hills_I4 | | | | Hills_I75 | | | |
|---|---|---|---|---|---|---|---|---|
| | w/o Micro | | w Micro | | w/o Micro | | w Micro | |
| | PI | MW | PI | MW | PI | MW | PI | MW |
| 5min | 92.5 | 14.8 | **90.9** | **13.2** | 92.3 | 11.7 | **90.5** | **10.5** |
| 10min | 93.4 | 17.6 | **90.6** | **14.9** | 91.6 | 13.1 | **89.7** | **11.2** |
| 15min | 92.9 | 18.7 | **89.7** | **15.6** | 93.1 | 14.7 | **89.6** | **11.5** |
| 35min | 93.1 | 22.2 | **89.1** | **16.2** | 93.0 | 17.1 | **89.5** | **12.1** |
| 45min | 93.9 | 25.5 | **89.1** | **16.5** | 94.0 | 19.1 | **88.9** | **12.5** |
| 60min | 93.5 | 27.1 | **88.5** | **17.0** | 93.6 | 19.3 | **89.9** | **12.8** |
| AVG | 93.2 | 21.0 | **89.7** | **16.0** | 92.9 | 15.8 | **89.7** | **12.0** |
| Predict time | Hills_I275 | | | | OS_I4 | | | |
| | w/o Micro | | w Micro | | w/o Micro | | w Micro | |
| | PI | MW | PI | MW | PI | MW | PI | MW |
| 5min | 92.8 | 15.3 | **90.8** | **13.6** | 90.5 | 11.0 | **91.0** | **11.0** |
| 10min | 92.8 | 17.5 | **90.4** | **15.8** | 91.0 | 12.8 | **90.8** | **12.4** |
| 15min | 93.0 | 19.0 | **90.1** | **16.3** | 91.2 | 14.0 | **90.4** | **12.6** |
| 35min | 92.6 | 21.2 | **89.3** | **17.1** | 91.4 | 15.9 | **89.4** | **13.3** |
| 45min | 92.6 | 23.4 | **88.4** | **17.3** | 91.4 | 16.7 | **89.2** | **13.9** |
| 60 min | 92.7 | 24.6 | **88.8** | **18.2** | 92.2 | 18.1 | **88.6** | **14.5** |
| AVG | 92.8 | 20.2 | **89.8** | **16.9** | 91.3 | 14.8 | **89.9** | **13.3** |

*:PI=PICP, MW=MPIW.

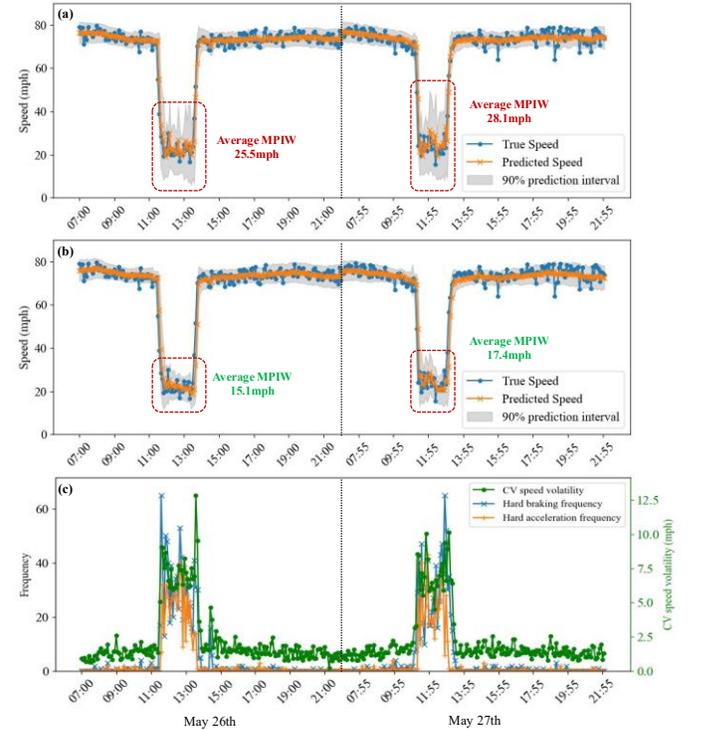

**Fig. 5.** Example of uncertainty prediction comparison (OS_I4, Seg_ID=104101).

## VI. DISCUSSIONS

*A. Cross-Attention Score Analysis*

To further demonstrate the effectiveness of cross-attention, **Fig. 6** presents detailed Cross-Attention (CA) scores under different traffic conditions. Taking a speed prediction sample as an example, **Fig. 6**(a) illustrates the macro and micro features (normalized to [0,1]) on Hills_I4 eastbound segments. Congested traffic is observed between segment IDs of 15-35 (speed <40mph). **Fig. 6**(b) visualizes the corresponding spatial CA scores, which quantify the relative impacts of micro features on macro features during speed forecasting. Results reveal that CA scores in congested segments are significantly higher than in free-flow segments, indicating that the model places greater importance on micro-behavior features (e.g., hard braking, CV speed volatility) in segment speed prediction. In contrast, segments under free-flow (e.g., segment IDs of 0-10) receive lower attention from micro features, indicating limited contributions on speed prediction. **Fig. 6**(c) displays that the distributions of CA scores are significantly different under different traffic conditions. In general, CA scores are low during free-flow and medium traffic but increase significantly under congested conditions. This suggests that micro driving behaviors (e.g., hard braking and acceleration) exert greater influence on traffic dynamics at lower speeds and during stop-and-go conditions on freeways. By contrast, in free-flow states, such behaviors are less frequent, prompting the model to focus more heavily on macro traffic features.

Overall, these findings highlight that, rather than applying fixed weights, the CA layer allows the model to adjust the influence of micro-level features according to the prevailing macro traffic state (e.g., free-flow vs. congestion). This adaptive mechanism enables MMCAformer to more effectively capture the bidirectional interactions between macro- and micro-level features, thereby achieving the precise and behavior-aware traffic speed prediction.

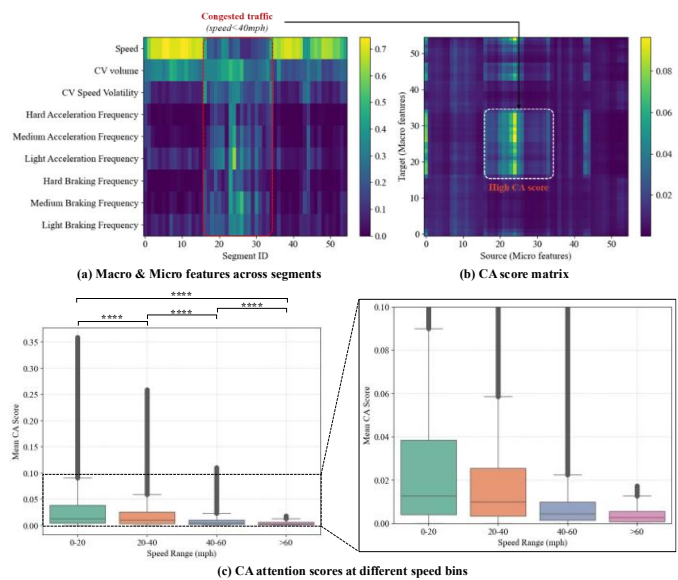

**Fig. 6.** Cross-Attention Score Visualization.

*B. Model Performances under Different CV Penetration*

Given that CV adoption is far from universal at the current stage, the model's performance under different CV penetration ratios (PRs) was investigated, as shown in **Fig. 7**. The figure illustrates the prediction errors at the four freeways under PRs of 1.0, 1.5, …, 4.0%. To emulate the lower penetration rate, unique trips in the CV dataset were randomly removed to reach the desired penetration rate. The average PR of raw Streetlight CV fleets is estimated at around 4%, which is used as the baseline for comparison.

From the results, it can be seen that the model performance declines slightly as the PR decreases from 4.0% to 1.0% on the four freeways. For example, on Hills_I4, the model MAE is 4.62mph at the lowest PR of 1.0%, improving to 4.54 mph at a PR of 1.5%. Finally, the MAE reduces to 4.46mph at the highest PR of 4.0% (baseline), representing a 3.4% decrease. Similar patterns are also observed on the other freeways. This degradation may be attributed to the reduced availability of critical micro driving behavior information (e.g., hard accelerations and braking) at lower PRs, which limits the model's ability to capture traffic dynamics for future speed evolution, thereby impairing predictive accuracy. Such an issue has also been reported in other CV-based prediction studies [18, 60]. Nevertheless, overall performance reductions remain within 3–8%, indicating that the model is relatively robust to variations in CV penetration rates.

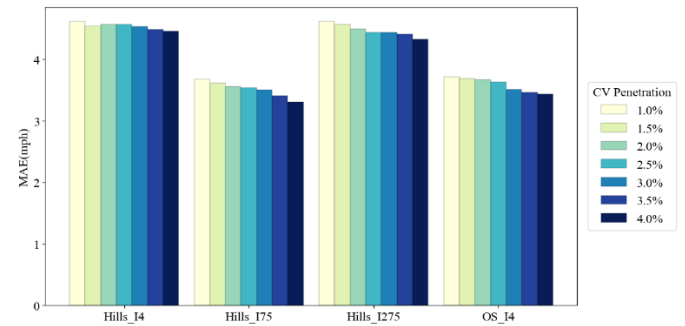

**Fig. 7.** Performances Under Different CV Penetrations.

*C. Prediction Error Distribution Analysis*

**Fig. 8** illustrates the distributions of prediction errors and the corresponding Q-Q plots under the Student-t distribution. The histograms on the left show that the errors are sharply concentrated around zero while exhibiting heavy tails on both sides, which are better captured by the Student-t distribution than the Gaussian distribution. To statistically validate this observation, we fitted the prediction errors with a Student-t distribution. The estimated degrees of freedom (df) across all freeways are below 3, confirming the presence of heavy-tailed error characteristics.

Furthermore, both the Kolmogorov–Smirnov (K-S) test and Q–Q plots were employed to support this assumption, as shown on the right side of **Fig. 8**. For the K-S test, the null hypothesis states that the errors follow a Student-t distribution; large test statistics with small p-values would lead to rejecting the hypothesis, whereas small statistics with large p-values would support it. Results across the four freeways consistently yield small K-S statistics (<0.05) with large p-values (>0.1), indicating that the null hypothesis cannot be rejected. For the Q–Q plots, the four plots demonstrate that the sample quantiles of errors align closely with the theoretical quantiles of the Student-t distribution (i.e., red line: y=x). These results provide strong evidence that the Student-t distribution is more appropriate to characterize the empirical error distribution than the conventional Gaussian assumption. This empirical validity confirms that adopting the Student-t Negative Log-likelihood Loss (t-NLL) is more suitable for speed prediction than the widely used Gaussian loss.

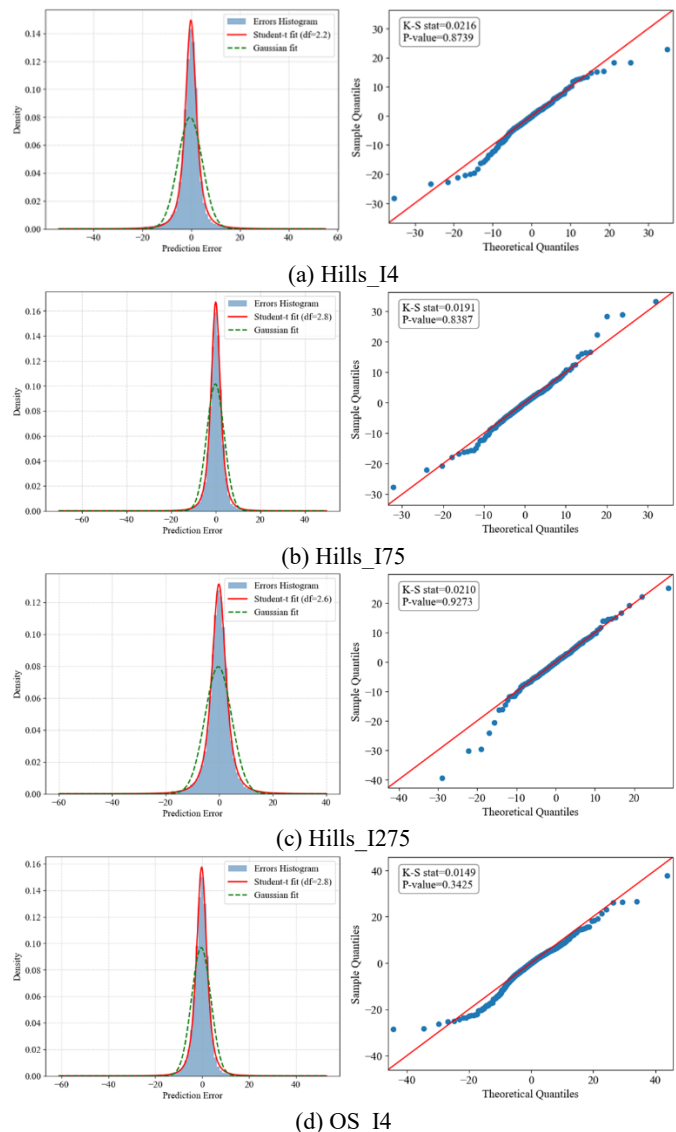

(a) Hills_I4

(b) Hills_I75

(c) Hills_I275

(d) OS_I4

**Fig. 8.** Prediction error distribution and the corresponding Q-Q plots under the Student-t distribution.

VII. CONCLUSION

In this study, we extracted micro driving behavior features from CV data and integrated them into traffic speed prediction. Unlike traditional methods that focus solely on macro traffic flow variables, we introduced a novel MMCAformer to capture spatiotemporal correlations between macro-level traffic status and micro-level driving behavior features. Optimized with the Student-t Negative Log-likelihood Loss, MMCAformer is designed to provide both point-wise predictions and predicted uncertainty. Experiments on Florida freeways demonstrate the benefits of introducing microscopic driving behavior features and the superior performance of our proposed MMCAformer. The main findings of the study can be summarized as:

1) Leveraging cross-attention to capture the spatiotemporal co-influences of macro and micro features, MMCAformer achieves the SOTA performance on speed prediction and reduces prediction errors by **6.7-12.8%** in RMSE, **5.9-11.9%** in MAE, and **3.5-18.8%** in MAPE.
2) Compared to solely using macro traffic flow information, introducing micro driving behavior features helps the model to capture critical traffic dynamics, thereby reducing RMSE, MAE, and MAPE by **9.0–14.0%, 6.9–10.1%,** and **10.2–20.5%**, respectively.
3) Integrating driving behavior features also helps to reduce model prediction uncertainty, as the MPIW decreased by **10.1-24.0 %** under a 90% significance level. Improvements are especially pronounced under congested conditions where risky driving maneuvers are most frequent.

The results prove the advantage of micro driving behavior features to enhance speed prediction accuracy and reliability. The proposed model can be used to implement real-time speed prediction and management in the ITS system. With the accurate speed prediction within the future time, it can help the ATM system predict potential traffic congestion areas, and target management measures can be pre-implemented to eliminate congestion. Meanwhile, by providing prediction uncertainty, traffic managers can better gauge the confidence level of each forecast and enhance the trustworthiness of traffic forecasts. For example, a tiered traffic response system can be established based on uncertainty-aware predictions. Low-uncertainty forecasts of congestion, which indicate urgent and highly probable events, could trigger automated and aggressive control measures. In contrast, high-uncertainty predictions may only prompt alerts for human operators, enabling them to monitor the situation more closely.

Nonetheless, there are still limitations for the current study. First, the used CV data has lower penetration at nighttime and suffers from significant data missing. Higher penetration or multiple data sources fusion could be considered in future research. Second, due to the data limitation, the proposed method was only tested in the four Florida freeways. More roadways should be tested in the future to validate model transferability and robustness. Last but not least, future research can explore combining more advanced modeling (e.g., generative Diffusion and Large Language model) to improve the prediction performance and the ability to capture complex traffic patterns.

ACKNOWLEDGMENT

This paper is funded partially by the Florida Department of Transportation. FDOT assumes no liability for the contents or use thereof.

REFERENCES

[1] A. Abdelraouf, M. Abdel-Aty, and N. Mahmoud, "Sequence-to-Sequence Recurrent Graph Convolutional Networks for Traffic Estimation and Prediction Using Connected Probe Vehicle Data," IEEE Transactions on Intelligent Transportation Systems, vol. 24, no. 1, pp. 1395–1405, Jan. 2023, doi: 10.1109/TITS.2022.3168865.

[2] W. Qian, T. D. Nielsen, Y. Zhao, K. G. Larsen, and J. J. Yu, "Uncertainty-Aware Temporal Graph Convolutional Network for Traffic Speed Forecasting," IEEE Transactions on Intelligent Transportation Systems, vol. 25, no. 8, pp. 8578–8590, Aug. 2024, doi: 10.1109/TITS.2024.3365721.

[3] H. Zhang, H. Dong, and Z. Yang, "TSGDiff: Traffic state generative diffusion model using multi-source information fusion," Transportation Research Part C: Emerging Technologies, vol. 174, p. 105081, May 2025, doi: 10.1016/j.trc.2025.105081.

[4] T. Mallick, J. Macfarlane, and P. Balaprakash, "Uncertainty Quantification for Traffic Forecasting Using Deep-Ensemble-Based Spatiotemporal Graph Neural Networks," IEEE Transactions on Intelligent Transportation Systems, vol. 25, no. 8, pp. 9141–9152, Aug. 2024, doi: 10.1109/TITS.2024.3381099.

[5] A. Sengupta, S. Mondal, A. Das, and S. I. Guler, "A Bayesian approach to quantifying uncertainties and improving generalizability in traffic prediction models," Transportation Research Part C: Emerging Technologies, vol. 162, p. 104585, May 2024, doi: 10.1016/j.trc.2024.104585.

[6] B. Yu, H. Yin, and Z. Zhu, "Spatio-Temporal Graph Convolutional Networks: A Deep Learning Framework for Traffic Forecasting," in Proceedings of the Twenty-Seventh International Joint Conference on Artificial Intelligence, July 2018, pp. 3634–3640. doi: 10.24963/ijcai.2018/505.

[7] Y. Zhang, Q. Zhou, J. Wang, A. Kouvelas, and M. A. Makridis, "CASAformer: Congestion-aware sparse attention transformer for traffic speed prediction," Communications in Transportation Research, vol. 5, p. 100174, Dec. 2025, doi: 10.1016/j.commtr.2025.100174.

[8] L. Zhao et al., "T-GCN: A Temporal Graph Convolutional Network for Traffic Prediction," IEEE Transactions on Intelligent Transportation Systems, vol. 21, no. 9, pp. 3848–3858, Sept. 2020, doi: 10.1109/TITS.2019.2935152.

[9] G. Zou, Z. Lai, C. Ma, Y. Li, and T. Wang, "A novel spatio-temporal generative inference network for predicting the long-term highway traffic speed," Transportation Research Part C: Emerging Technologies, vol. 154, p. 104263, Sept. 2023, doi: 10.1016/j.trc.2023.104263.

[10] Y. Zou, Y. Chen, Y. Xu, H. Zhang, and S. Zhang, "Short-term freeway traffic speed multistep prediction using an iTransformer model," Physica A: Statistical Mechanics and its Applications, vol. 655, p. 130185, Dec. 2024, doi: 10.1016/j.physa.2024.130185.

[11] Y. Gao and D. Levinson, "Lane changing and congestion are mutually reinforcing?," Communications in Transportation Research, vol. 3, p. 100101, Dec. 2023, doi: 10.1016/j.commtr.2023.100101.

[12] S. Heshami and L. Kattan, "A stochastic microscopic based freeway traffic state and spatial-temporal pattern prediction in a connected vehicle environment," Journal of Intelligent Transportation Systems, vol. 28, no. 3, pp. 313–339, May 2024, doi: 10.1080/15472450.2022.2130291.

[13] S.-E. Molzahn, B. S. Kerner, and H. Rehborn, "Phase based jam warnings: an analysis of synchronized flow with floating car data," Journal of Intelligent Transportation Systems, vol. 24, no. 6, pp. 569–584, Nov. 2020, doi: 10.1080/15472450.2019.1638781.

[14] G. Li, Knoop, Victor L., and H. and van Lint, "How predictable are macroscopic traffic states: a perspective of uncertainty quantification," Transportmetrica B: Transport Dynamics, vol. 12, no. 1, p. 2314766, Dec. 2024, doi: 10.1080/21680566.2024.2314766.

[15] M. Treiber and A. Kesting, Traffic Flow Dynamics: Data, Models and Simulation. Berlin, Heidelberg: Springer Berlin Heidelberg, 2013. doi: 10.1007/978-3-642-32460-4.

[16] H. Yao, Q. Li, and X. Li, "A study of relationships in traffic oscillation features based on field experiments," Transportation Research Part A: Policy and Practice, vol. 141, pp. 339–355, Nov. 2020, doi: 10.1016/j.tra.2020.09.006.

[17] Z. Zheng, S. Ahn, D. Chen, and J. Laval, "Freeway Traffic Oscillations: Microscopic Analysis of Formations and Propagations using Wavelet Transform," Procedia - Social and Behavioral Sciences, vol. 17, pp. 702–716, Jan. 2011, doi: 10.1016/j.sbspro.2011.04.540.

[18] L. Han, M. Abdel-Aty, R. Yu, and C. Wang, "LSTM + Transformer Real-Time Crash Risk Evaluation Using Traffic Flow and Risky Driving Behavior Data," IEEE Transactions on Intelligent Transportation Systems, pp. 1–13, 2024, doi: 10.1109/TITS.2024.3438616.

[19] X. Wang et al., "Traffic light optimization with low penetration rate vehicle trajectory data," Nat Commun, vol. 15, no. 1, p. 1306, Feb. 2024, doi: 10.1038/s41467-024-45427-4.

[20] Y. Gu, D. Liu, R. Arvin, A. J. Khattak, and L. D. Han, "Predicting intersection crash frequency using connected vehicle data: A framework for geographical random forest," Accident Analysis & Prevention, vol. 179, p. 106880, Jan. 2023, doi: 10.1016/j.aap.2022.106880.

[21] L. Han, R. Yu, C. Wang, and M. Abdel-Aty, "Transformer-based modeling of abnormal driving events for freeway crash risk evaluation," Transportation Research Part C: Emerging Technologies, vol. 165, p. 104727, Aug. 2024, doi: 10.1016/j.trc.2024.104727.

[22] M. Kamrani, B. Wali, and A. J. Khattak, "Can Data Generated by Connected Vehicles Enhance Safety?: Proactive Approach to Intersection Safety Management," Transportation Research Record, vol. 2659, no. 1, pp. 80–90, Jan. 2017, doi: 10.3141/2659-09.

[23] S. Zhang and M. Abdel-Aty, "Real-time crash potential prediction on freeways using connected vehicle data," Analytic Methods in Accident Research, vol. 36, p. 100239, Dec. 2022, doi: 10.1016/j.amar.2022.100239.

[24] R. Arvin, M. Kamrani, and A. J. Khattak, "How instantaneous driving behavior contributes to crashes at intersections: Extracting useful information from connected vehicle message data," Accident Analysis & Prevention, vol. 127, pp. 118–133, June 2019, doi: 10.1016/j.aap.2019.01.014.

[25] R. Cheng, M. Liu, and Y. Xu, "ST_AGCNT: Traffic Speed Forecasting Based on Spatial–Temporal Adaptive Graph Convolutional Network with Transformer," Sustainability, vol. 17, no. 5, Art. no. 5, Jan. 2025, doi: 10.3390/su17051829.

[26] C. Song, Y. Lin, S. Guo, and H. Wan, "Spatial-Temporal Synchronous Graph Convolutional Networks: A New Framework for Spatial-Temporal Network Data Forecasting," Proceedings of the AAAI Conference on Artificial Intelligence, vol. 34, no. 01, Art. no. 01, Apr. 2020, doi: 10.1609/aaai.v34i01.5438.

[27] N. Ouyang et al., "Graph Transformer-Based Dynamic Edge Interaction Encoding for Traffic Prediction," IEEE Transactions on Intelligent Transportation Systems, vol. 26, no. 3, pp. 4066–4079, Mar. 2025, doi: 10.1109/TITS.2024.3513325.

[28] Y. Xie, Y. Xiong, and Y. Zhu, "ISTD-GCN: Iterative Spatial-Temporal Diffusion Graph Convolutional Network for Traffic Speed Forecasting," Aug. 10, 2020, arXiv: arXiv:2008.03970. doi: 10.48550/arXiv.2008.03970.

[29] A. Abdelraouf, M. Abdel-Aty, and J. Yuan, "Utilizing Attention-Based Multi-Encoder-Decoder Neural Networks for Freeway Traffic Speed Prediction," IEEE Transactions on Intelligent Transportation Systems, vol. 23, no. 8, pp. 11960–11969, Aug. 2022, doi: 10.1109/TITS.2021.3108939.

[30] S. Yu et al., "A traffic state prediction method based on spatial–temporal data mining of floating car data by using autoformer architecture," Computer-Aided Civil and Infrastructure Engineering, vol. 39, no. 18, pp. 2774–2787, 2024, doi: 10.1111/mice.13179.

[31] A. A. Awan, A. Majid, R. Riaz, S. S. Rizvi, and S. J. Kwon, "A Novel Deep Stacking-Based Ensemble Approach for Short-Term Traffic Speed Prediction," IEEE Access, vol. 12, pp. 15222–15235, 2024, doi: 10.1109/ACCESS.2024.3357749.

[32] L. Chang, C. Ma, K. Sun, Z. Qu, and C. Ren, "Enhanced road information representation in graph recurrent network for traffic speed prediction," IET Intelligent Transport Systems, vol. 17, no. 7, pp. 1434–1453, 2023, doi: 10.1049/itr2.12334.

[33] D. Chen, S. Ahn, J. Laval, and Z. Zheng, "On the periodicity of traffic oscillations and capacity drop: The role of driver characteristics," Transportation Research Part B: Methodological, vol. 59, pp. 117–136, Jan. 2014, doi: 10.1016/j.trb.2013.11.005.

[34] B. Wali, A. J. Khattak, H. Bozdogan, and M. Kamrani, "How is driving volatility related to intersection safety? A Bayesian heterogeneity-based

analysis of instrumented vehicles data," Transportation Research Part C: Emerging Technologies, vol. 92, pp. 504–524, July 2018, doi: 10.1016/j.trc.2018.05.017.

[35] Z. Islam and M. Abdel-Aty, "Traffic conflict prediction using connected vehicle data," Analytic Methods in Accident Research, vol. 39, p. 100275, Sept. 2023, doi: 10.1016/j.amar.2023.100275.

[36] L. Liu, Z. Qiu, G. Li, Q. Wang, W. Ouyang, and L. Lin, "Contextualized Spatial–Temporal Network for Taxi Origin-Destination Demand Prediction," IEEE Transactions on Intelligent Transportation Systems, vol. 20, no. 10, pp. 3875–3887, Oct. 2019, doi: 10.1109/TITS.2019.2915525.

[37] C. M. J. Tampere and L. H. Immers, "An Extended Kalman Filter Application for Traffic State Estimation Using CTM with Implicit Mode Switching and Dynamic Parameters," in 2007 IEEE Intelligent Transportation Systems Conference, Sept. 2007, pp. 209–216. doi: 10.1109/ITSC.2007.4357755.

[38] M. Lippi, M. Bertini, and P. Frasconi, "Short-Term Traffic Flow Forecasting: An Experimental Comparison of Time-Series Analysis and Supervised Learning," IEEE Transactions on Intelligent Transportation Systems, vol. 14, no. 2, pp. 871–882, June 2013, doi: 10.1109/TITS.2013.2247040.

[39] H. Wang, Liu ,Lu, Dong ,Shangjia, Qian ,Zhen, and H. and Wei, "A novel work zone short-term vehicle-type specific traffic speed prediction model through the hybrid EMD–ARIMA framework," Transportmetrica B: Transport Dynamics, vol. 4, no. 3, pp. 159–186, Sept. 2016, doi: 10.1080/21680566.2015.1060582.

[40] B. Yao et al., "Short-Term Traffic Speed Prediction for an Urban Corridor," Computer-Aided Civil and Infrastructure Engineering, vol. 32, no. 2, pp. 154–169, 2017, doi: 10.1111/mice.12221.

[41] W. Wu, Y. Xia, and W. Jin, "Predicting Bus Passenger Flow and Prioritizing Influential Factors Using Multi-Source Data: Scaled Stacking Gradient Boosting Decision Trees," IEEE Transactions on Intelligent Transportation Systems, vol. 22, no. 4, pp. 2510–2523, Apr. 2021, doi: 10.1109/TITS.2020.3035647.

[42] B. Jiang and Y. Fei, "Vehicle Speed Prediction by Two-Level Data Driven Models in Vehicular Networks," IEEE Transactions on Intelligent Transportation Systems, vol. 18, no. 7, pp. 1793–1801, July 2017, doi: 10.1109/TITS.2016.2620496.

[43] A. Csikós, Z. J. Viharos, K. B. Kis, T. Tettamanti, and I. Varga, "Traffic speed prediction method for urban networks — an ANN approach," in 2015 International Conference on Models and Technologies for Intelligent Transportation Systems (MT-ITS), June 2015, pp. 102–108. doi: 10.1109/MTITS.2015.7223243.

[44] Z. Zhou, Z. Yang, Y. Zhang, Y. Huang, H. Chen, and Z. Yu, "A comprehensive study of speed prediction in transportation system: From vehicle to traffic," iScience, vol. 25, no. 3, p. 103909, Mar. 2022, doi: 10.1016/j.isci.2022.103909.

[45] Y. Li, R. Yu, C. Shahabi, and Y. Liu, "Diffusion Convolutional Recurrent Neural Network: Data-Driven Traffic Forecasting," Feb. 22, 2018, arXiv: arXiv:1707.01926. doi: 10.48550/arXiv.1707.01926.

[46] Z. Wu, S. Pan, G. Long, J. Jiang, and C. Zhang, "Graph wavenet for deep spatial-temporal graph modeling," in Proceedings of the 28th International Joint Conference on Artificial Intelligence, in IJCAI'19. Macao, China: AAAI Press, Aug. 2019, pp. 1907–1913.

[47] S. Guo, Y. Lin, N. Feng, C. Song, and H. Wan, "Attention Based Spatial-Temporal Graph Convolutional Networks for Traffic Flow Forecasting," AAAI, vol. 33, no. 01, pp. 922–929, July 2019, doi: 10.1609/aaai.v33i01.3301922.

[48] C. Zheng, X. Fan, C. Wang, and J. Qi, "GMAN: A Graph Multi-Attention Network for Traffic Prediction," AAAI, vol. 34, no. 01, pp. 1234–1241, Apr. 2020, doi: 10.1609/aaai.v34i01.5477.

[49] H. Li, J. Liu, S. Han, J. Zhou, T. Zhang, and C. L. Philip Chen, "STFGCN: Spatial–temporal fusion graph convolutional network for traffic prediction," Expert Systems with Applications, vol. 255, p. 124648, Dec. 2024, doi: 10.1016/j.eswa.2024.124648.

[50] H. Liu et al., "Spatio-Temporal Adaptive Embedding Makes Vanilla Transformer SOTA for Traffic Forecasting," in Proceedings of the 32nd ACM International Conference on Information and Knowledge Management, Birmingham United Kingdom: ACM, Oct. 2023, pp. 4125–4129. doi: 10.1145/3583780.3615160.

[51] J. Zhang, Y. Chen, T. Wang, C.-Z. T. Xie, and Y. Tian, "Mixture of Spatial–Temporal Graph Transformer Networks for urban congestion prediction using multimodal transportation data," Expert Systems with Applications, vol. 268, p. 126108, Apr. 2025, doi: 10.1016/j.eswa.2024.126108.

[52] S. Martin and A. Bigazzi, "Cyclist Perception–Reaction Time and Stopping Sight Distance for Unexpected Hazards," Journal of Transportation Engineering, Part A: Systems, vol. 151, no. 6, p. 04025030, June 2025, doi: 10.1061/JTEPBS.TEENG-8805.

[53] Q. Packer, Ivan ,John, and M. and Filipovska, "Hard braking collected from probe vehicles as a predictor of crash occurrence at urban intersections in Connecticut," Journal of Transportation Safety & Security, vol. 0, no. 0, pp. 1–31, doi: 10.1080/19439962.2025.2476993.

[54] Y. Wang et al., "TimeXer: Empowering Transformers for Time Series Forecasting with Exogenous Variables," Nov. 11, 2024, arXiv: arXiv:2402.19072. doi: 10.48550/arXiv.2402.19072.

[55] A. Kumar, T. K. Marks, W. Mou, C. Feng, and X. Liu, "UGLLI Face Alignment: Estimating Uncertainty with Gaussian Log-Likelihood Loss," in 2019 IEEE/CVF International Conference on Computer Vision Workshop (ICCVW), Seoul, Korea (South): IEEE, Oct. 2019, pp. 778–782. doi: 10.1109/ICCVW.2019.00103.

[56] D. A. Nix and A. S. Weigend, "Estimating the mean and variance of the target probability distribution," in Proceedings of 1994 IEEE International Conference on Neural Networks (ICNN'94), June 1994, pp. 55–60 vol.1. doi: 10.1109/ICNN.1994.374138.

[57] A. Kendall and Y. Gal, "What Uncertainties Do We Need in Bayesian Deep Learning for Computer Vision?," in Advances in Neural Information Processing Systems, Curran Associates, Inc., 2017. Accessed: June 26, 2025. [Online]. Available: https://proceedings.neurips.cc/paper/2017/hash/2650d6089a6d640c5e85b2b88265dc2b-Abstract.html

[58] Y. Liu et al., "SAFER-predictor: Sparse adversarial training framework for robust traffic prediction under missing and noisy data," Communications in Transportation Research, vol. 5, p. 100192, Dec. 2025, doi: 10.1016/j.commtr.2025.100192.

[59] A. M. Avila and I. Mezić, "Data-driven analysis and forecasting of highway traffic dynamics," Nat Commun, vol. 11, no. 1, Art. no. 1, Apr. 2020, doi: 10.1038/s41467-020-15582-5.

[60] N. Gupta, H. Jashami, P. T. Savolainen, T. J. Gates, T. Barrette, and W. Powell, "Examining the Relationship between Connected Vehicle Driving Event Data and Police-Reported Traffic Crash Data at the Segment- and Event Level," Transportation Research Record, p. 03611981241243329, Apr. 2024, doi: 10.1177/03611981241243329.

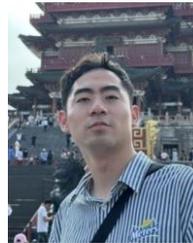

**Lei Han** received the bachelor's degree and master's degree in the College of Transportation Engineering, Tongji University in 2020 and 2023, respectively. He is currently pursuing a Ph.D. degree in transportation engineering from the University of Central Florida (UCF). He is currently a research associate with UCF. His research interests include traffic safety analysis, connected vehicle data applications, intelligent transportation systems, and deep learning applications in transportation engineering.

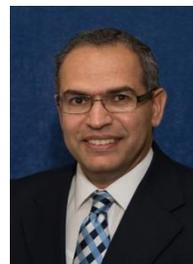

**Mohamed Abdel-Aty** (Senior member, IEEE) is a Pegasus Professor and Trustee Chair at UCF, Orlando, FL, USA. He is leading the Future City initiative at the UCF. He is also the Director of the Smart and Safe Transportation Laboratory. He has managed over 90 research projects. He has delivered more than 35 keynote speeches at conferences around the world. He has published more than 480 journals (As of September 2025, Google Scholar citations: over 39800, H-index: 106). His main expertise and interests are in the areas of ITS, simulation, CAV, and active traffic management. He is the Editor Emeritus of Accident Analysis and Prevention. Dr. Aty has received the 2020 Roy Crum Distinguished Service Award from the Transportation Research Board, National Safety Council's Distinguished Service to Safety Award, Francis Turner award from ASCE and the Lifetime Achievement Safety Award and S.S. Steinberg Award from ARTBA in 2019 and 2022, respectively. He has also received with his team multiple international awards including the Prince Michael Road Safety Award, London 2019.

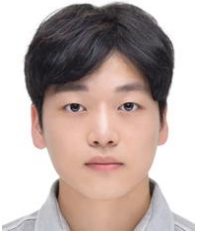

**Younggun Kim** received the B.S. degree in mechanical engineering from Ajou University, Suwon, South Korea, in 2024. He is currently pursuing the M.S. degree in civil engineering with the University of Central Florida, FL, USA. His research interests include artificial intelligence, intelligent transportation systems, and autonomous vehicles.

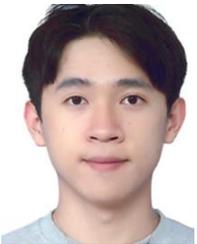

**Yang-Jun Joo** received the Ph.D. degree in civil and environmental engineering from Seoul National University, Seoul, South Korea, in 2023. He is currently a Post-Doctoral Researcher with the Department of Civil and Environmental Engineering, University of Central Florida. His research interests include intelligent transportation systems, driving risk assessment, and traffic safety.

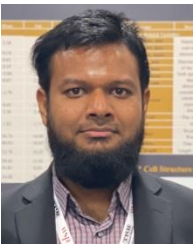

**Zubayer Islam** (member, IEEE) received the master's degree in electrical and electronics engineering from the Bangladesh University of Engineering and Technology, Dhaka, Bangladesh, in 2017, and the Ph.D. degree in transportation engineering from the University of Central Florida, Orlando, FL, USA, in 2021, where he is currently an Assistant Professor of Transportation Engineering.